\documentclass[a4paper,english]{rnti}  

\usepackage[T1]{fontenc}
\usepackage[latin1]{inputenc}
\usepackage{numprint}
\usepackage{enumitem}

\usepackage{amsmath,amssymb}
\usepackage{siunitx}

\usepackage{url}

\usepackage{graphicx}
\usepackage{xcolor}

\titrecourt{Fine-tuning BERT-based models for Plant Health Bulletin Classification}

\nomcourt{I. NomPremierAuteur et al.}

\titre{Fine-tuning BERT-based models for Plant Health Bulletin Classification}

\auteur{Shufan Jiang\affil{1}\affilsep\affil{2},
        Rafael Angarita\affil{1},\\
        St\'ephane Cormier\affil{2},\
        Francis Rousseaux\affil{2}}

\affiliation{
    \affil{1}Institut Sup\'erieur d'Electronique de Paris, LISITE, Paris, France\\
          name.lastname@isep.fr,\\
    \affil{2} Universit\'e de Reims Champagne Ardenne, CReSTIC, Reims, France\\
          name.lastname@univ-reims.fr\\
 }

\resume{%
In the era of digitization, different actors in agriculture produce numerous data. Such data contains already latent historical knowledge in the domain. This knowledge enables us to precisely study natural hazards within global or local aspects, and then improve the risk prevention tasks and augment the yield, which helps to tackle the challenge of growing population and changing alimentary habits. In particular, French Plants Health Bulletins (BSV, for its name in French Bulletin de Sant\'e du V\'eg\'etal) give information about the development stages of phytosanitary risks in agricultural production. However, they are written in natural language, thus,  machines and human cannot exploit them as efficiently as it could be. Natural language processing (NLP) technologies aim to automatically process and analyze large amounts of natural language data. Since the 2010s, with the increases in computational power and parallelization, representation learning and deep learning methods became widespread in NLP. Recent advancements Bidirectional Encoder Representations from Transformers (BERT) inspire us to rethink of knowledge representation and natural language understanding in plant health management domain. The goal in this work is to propose a BERT-based approach to automatically classify the BSV to make their data easily indexable. We sampled 200 BSV to finetune the pretrained BERT language models and classify them as pest or/and disease and we show preliminary results.
}

\begin{document}

\section{Introduction}
In the era of digitization, different actors in agriculture produce numerous data. Such data contains already latent historical knowledge in the domain. This knowledge enables us to precisely study natural hazards within global or local aspects, and then improve the risk prevention tasks and augment the yield, which helps to tackle the challenge of growing population and changing alimentary habits. In particular, French Plants Health Bulletins (BSV, for its name in French Bulletin de Sant\'e du V\'eg\'etal) give information about the development stages of phytosanitary risks in agricultural production~\cite{hal-02536389}. BSV are published periodically by the French Regional Plant Protection Services (SRPV) and Groupings Protection against Harmful Organisms (GDON). These data is collected by the observation network Epiphyt \footnote{https://agroedieurope.fr/wp-content/uploads/fiche-projet-epiphyt-fr.pdf} to build a national database. These data are collected following agronomic observations made throughout France by regional monitoring networks involving 400 observers in 36 partner networks.

Following the \textit{social sensing} paradigm~\cite{wang2015social}, individuals -whether they are farmers or not- have more and more connectivity to information while on the move, at the field-level. Each individual can become a broadcaster of information. In this sense, 
less formal than the BSV but relevant and usually real-time hazard information is also published in social networks such as Twitter. 
Given the nature of such publications, it is not straightforward to efficiently and effectively take advantage of the information they contain, let alone doing it automatically and relating these data to data coming other sources such as sensors or other information systems.
To handle, process and make these data searchable, it is necessary to start by classifying its textual content automatically.

Text classification is a category of Natural Language
Processing (NLP), which employs computational techniques for the purpose of learning,
understanding, and producing human language content~\cite{hirschberg2015advances}.
A well-known feature representation technique for text classification is Word2Vec~\cite{goldberg2014word2vec} and some real-world applications of text classification are spam identification and fraud and bot detection~\cite{howard2018universal}.
PestObserver proposedto index BSV with crops, bioagressors and diseases based on concurrence analysis, regex and pattern matching text classification techniques ~\cite{turenne:hal-01145489}. However, the tags are not complete: some bulletins are only indexed with crop while the content do mention bioagressors or diseases. User-defined rules were used for relation extraction. This is an efficient and precise approach, but it cannot represent the latent semantic relations. Human annotation was also needed to enrich the project annotation resources and dictionaries. In overall, the global PestObserver approach relies on highly crowdsourcing-dependent techniques, which made the information extraction procedure not dynamic enough to adapt to changes in document format or contents.

Recent advancements in Bidirectional Encoder Representations from Transformers (BERT)~\cite{devlin2019bert} have showed important improvements in NLP; for example, the efficiency of fine-tuned BERT models for Multi-Label tweets classification has been proved in disaster monitoring \cite{Zahera2019FinetunedBM}.
We find it also important to include observation from non-expert individuals which are not necessarily in the network of experts collecting data. Information about such observation may be extracted from social media like twitter. Thus, our goal is to build a general language model that represents the phytosanitary risks to improve the information extraction and document classification from heterogeneous textual data sources. 
The BSV might serve as a corpus to train this language model.

In this work, we aim to explore the potentials and limits of BERT considering the available data sets. More precisely, we want to answer the following questions:
Will BERT be able to give an interesting classification for BSV compared to PestObeserver? And how well can it generalize for natural hazard prediction from heterogeneous documents?

\section{Data}

\textit{Existing Dataset}
    \begin{enumerate}[label=\Alph*)]
        \item \label{item:A} We downloaded BSVs~\cite{BSV:OCR} from the PestOberserver site. In this collection of 40828 files, there are 17286 older BSVs in XML format, and 23542 OCR (Optical Character Recognition) processed BSVs in plain-text format.
        \item \label{item:B} We also obtained tags for each BSV from the PestOberserver site. There are 389 \textit{bioagressor} and 279 \textit{disease} tags and those BSVs were annotated using text mining techniques and by domain experts.  Unfortunately,  only the plain-text files are annotated as \textit{bioagressors} or \textit{diseases}. The XML files are annotated only with crops names.
        \item \label{item:C} We use concepts in FrenchCropUsuage thesaurus ~\cite{https://doi.org/10.25504/fairsharing.9228fv} and the tags in item~\ref{item:B} as filters to collect tweets for testing the classification model. 
    \end{enumerate}

\textit{Linguistic Prepossessing for text of each BSV }\\
    We removed the following from the text of each BSV:
    \begin{itemize}
        \item Punctuation marks, URLs, phone numbers and stop words from the BSV text.
        \item Extra white-spaces, repeated full stops, question marks and exclamation marks.
        \item Continuous lines that contain less than 3 words are rows from broken tables in the original PDF file. 
        \item Strings like "B U L L E T I N" appearing in vertical lines. 
    \end{itemize}

\textit{Dataset Construction}
    \begin{itemize}
        \item For the unsupervised fine-tuning task, we extracted paragraphs from xml format BSV in item~\ref{item:A} to make the corpus for the self-supervised fine-tuning of the language model.
        \item For the classification of the topic, we randomly split 200 cleaned BSVs into 4000 chunks containing between 5 and 256 words. We classify each chuck as \textit{bioagressors} and \textit{diseases} according to the tags of its corresponding BSV -see item~\ref{item:C}-.
        \item We also manually classified 400 sentences extracted from cleaned BSVs. We classified these sentences as \textit{bioagressor} and \textit{disease} if the BSV says the threshold of danger is reached, or if it recommends to apply a treatment. This classification task aims to test if the language model can ``understand'' the risk.
    \end{itemize}

\section{Training details}
All the experiments were conducted on a workstation having Intel Core i9-9900K CPU, 32GB memory, 1 single NVIDIA TITAN RTX GPU with CUDA 10.0.130, trasformers\cite{wolf-etal-2020-transformers} and fast-bert\cite{fast-bert}. 
For the multi-label classification of the topic, we tested respectively the pretrained CamemBERT model\cite{Martin2020CamemBERTAT} and BERT-Base, Multilingual Cased model \cite{MultilingualBERT}. 
We first fine-tune the language model to adapt it to domain specific context. All hyper-parameters are tuned on raw BSV corpus over 2 epochs as suggested by the author \cite{devlin2019bert}.  The batch size is 8. Adam ~\cite{kingma2017adam}is used for optimization with an initial learning rate of 1e-4.
Then we used these fine-tuned language models to train the classification task. The batch size is 8. Max sequence length is set to 256.  Adam\cite{kingma2017adam}is used for optimization with an initial learning rate of 2e-5. We trained the classification model for 5 or 10 epochs and saved the one with better f1 score. The model's output are the probabilities of all classes, we hence set threshold value of 0.5 to pickup a list
of possible classes to the input text as final prediction.
 Finally we evaluate the model over the test set, we also test the model with tweets that talks about natural hazards.
For the multi-label classification of risk, the goal is to predict the presence of bioagressor risk or disease risk, the experiment setup is the same as the previous task, except that we fine-tuned the models on raw BSV corpus and on paragraphs extracted from BSV in XML format -see item~\ref{item:A}-.

\section{Results}

To evaluate all these classifications, we use accuracy, precision, recall, F1 score and ROC\_AUC score \cite{Hossin2015}. 

Table~\ref{tab:topic-camemBERT} shows results of multi-label classification (bioagressor and disease) task. As the dataset is simply tagged with the appearance of key words, moreover, the tags on pestobserver site are not completed, the pertinence of its categorisation is limited. However, we observed that the model can correct some of false negative taggings from pestobserver (which means, a phrase which mentions borer and which not tagged as bioagressor on the pestobserver site, may still be classified as bioagressor by our model). This model also shows certain generalizability when tested with tweets item~\ref{item:C}, of which the syntax is unknown to the model. As an example, consider the following text about ``pyrale'' (pyralid moths) from a BSV:

\begin{quote} 
\centering 
\textit{``Dans les pi\`eges lumineux, le nombre de captures correspond \`a la fois aux individus m\^ales et femelles. Cartographie des captures des pyrales dans les pi\`eges \`a ph\'eromone dans les Pays de la Loire (L\'egende : vert : absence, orange : 1-4 pyrales, rouge : 5 et + pyrales).''}
\end{quote}

For the previous example paragraph, PestObserver has no tag for it; however, our classifier predicts it to be bioagressor.

\begin{table}[ht]
    \centering
    \caption{prediction of the topic (threshold=0.5)using CamemBERT model}
    \npdecimalsign{.}
    \nprounddigits{2}
    \begin{tabular}[t]{l n{2}{2} n{2}{2} n{2}{2} n{2}{2} n{2}{2}}
        \hline
         & {Accuracy} & {Precision} & {Recall} & {F Score} & {ROC\_AUC}\\
        \hline
        Bioagressor & 0.8558139534883721 & 0.76454294 &0.87619048 & 0.81656805 & \\
        Disease & 0.8953488372093024 & 0.68609865 & 0.88439306 & 0.77272727 & \\
        Weighted Average & & 0.736733795482217 & 0.8790983606557377 & 0.8010261333874197 & 0.9113796040025547\\
        \hline
    \end{tabular}
    \npnoround
    \label{tab:topic-camemBERT}
\end{table}

\begin{table}[ht]
    \centering
    \caption{prediction of the topic (threshold=0.5) using BERT-Base, Multilingual Cased model }
    \npdecimalsign{.}
    \nprounddigits{2}
    \begin{tabular}[t]{l n{2}{2} n{2}{2} n{2}{2} n{2}{2} n{2}{2}}
        \hline
        &{Accuracy} &{Precision} &{Recall} &{F Score}&{ROC\_AUC}\\
        \hline
        Bioagressor &0.8662790697674418 &0.78409091 &0.87619048 &0.82758621 & \\
        Disease &0.8965116279069767 &0.69585253 &0.87283237 &0.77435897 & \\
        Weighted Average & &0.7528096820551209 &0.875 &0.8087167166731892 &0.9105695789865872\\
        \hline
    \end{tabular}
    \npnoround
    \label{tab:topic-BERT-Multilingual-cased}
\end{table}

Table~\ref{tab:topic-BERT-Multilingual-cased} shows the results of the same multi-label classification task with BERT-Base, Multilingual Cased model. The scores are slightly better than the ones produced by CamemBERT presented in Table~\ref{tab:topic-camemBERT}; however, the size of the pre-trained BERT-Base, Multilingual Cased model is bigger than CamemBERT -since it covers more than 104 languages- and it takes more time for the training.

Table~\ref{tab:risk} shows the results of multi-label classification of risks task. In this experiment, the pertinence of the training set is assured by manual annotation. We noticed that the language model is able to detect the presences of bioagressor or disease in the text though the dataset is much smaller than the one of previous classification task, which is equivalent to filter the document with a given domain key words list. Considering the risk level or the detection of the positive/negative sense of the phrase, the prediction is less pertinent.  For example, phrases like the following are still classified to having a risk of bioagressor even tough it says there is only a small presence of bioagressor so no action is required. These results may be improved if more data is available.

\begin{quote} 
\centering 
\textit{``... note l'apparition des premiers pucerons \`a villenauxe la petite (77) avec moins de 1 puceron par feuille.
le seuil d'intervention, de 5 \`a 10 pucerons par feuille, n'est pas encore atteint. aucune intervention n'est justifi\'ee.''}
\end{quote}

\begin{table}[ht]
    \centering
    \caption{prediction of risks:}
    \npdecimalsign{.}
    \nprounddigits{2}
    \begin{tabular}[t]{l n{2}{2} n{2}{2} n{2}{2} n{2}{2} n{2}{2}}
        \hline
        &{Accuracy} &{Precision} &{Recall} &{F Score} &{ROC\_AUC}\\
        \hline
        Bioagressor &0.85 &0.62962963 &0.89473684 &0.73913043 & \\
        Disease &0.825 &0.72222222 &0.59090909 & 0.65 & \\
        Weighted Average & &0.6793134598012648 &0.7317073170731707 &0.65  &0.9135069592292953\\
        \hline
    \end{tabular}
    \npnoround
    \label{tab:risk}
\end{table}

\section{Conclusion}

Recent advancements BERT are promising regarding natural language processing.
Our objective is to classify agricultural-related documents according to natural hazards they discuss.
We have studied existing textual data in french plant health domain, specially the BSVs, and experimented with the BERT multilingual and CamemBERT models.
Our results show that fine-tuned BERT-based model is sufficient for the topic prediction of BSV.
The preliminary prediction test on tweets convinced us that BERT-based models is generalizable for representing features in the French plant health domain. 
For our future work, we plan to feed our model with more pertinent data.
It may be also interesting to explore alternatives such as FlauBERT\cite{le2020flaubert}, which another BERT-based language model for French. Finally, We also plan to investigate feature-based approaches with BERT embeddings.

\bibliographystyle{rnti}
\bibliography{bibliographie}

\providecommand\Fr{}
\providecommand\Eng{}
\providecommand\andname{and}
\providecommand\andnamec{and}

\begin{thebibliography}{}


\bibitem[{Devlin et~al.}(2019){Devlin, Chang, Lee, \andnamec{}
  Toutanova}]{devlin2019bert}
Devlin, J., M.-W. Chang, K.~Lee, \andname{} K.~Toutanova (2019).
\newblock Bert: Pre-training of deep bidirectional transformers for language
  understanding.

\bibitem[{{FAIRsharing Team}}(2018){{FAIRsharing
  Team}}]{https://doi.org/10.25504/fairsharing.9228fv}
{FAIRsharing Team} (2018).
\newblock Fairsharing record for: French crop usage.

\bibitem[{Goldberg \andnamec{} Levy}(2014){Goldberg \andnamec{}
  Levy}]{goldberg2014word2vec}
Goldberg, Y. \andname{} O.~Levy (2014).
\newblock word2vec explained: deriving mikolov et al.'s negative-sampling
  word-embedding method.
\newblock {\em arXiv preprint arXiv:1402.3722\/}.

\bibitem[{Hirschberg \andnamec{} Manning}(2015){Hirschberg \andnamec{}
  Manning}]{hirschberg2015advances}
Hirschberg, J. \andname{} C.~D. Manning (2015).
\newblock Advances in natural language processing.
\newblock {\em Science\/}~{\em 349\/}(6245), 261--266.

\bibitem[{Hossin \andnamec{} M.N}(2015){Hossin \andnamec{} M.N}]{Hossin2015}
Hossin, M. \andname{} S.~M.N (2015).
\newblock A review on evaluation metrics for data classification evaluations.
\newblock {\em International Journal of Data Mining \& Knowledge Management
  Process\/}~{\em 5}, 01--11.

\bibitem[{Howard \andnamec{} Ruder}(2018){Howard \andnamec{}
  Ruder}]{howard2018universal}
Howard, J. \andname{} S.~Ruder (2018).
\newblock Universal language model fine-tuning for text classification.

\bibitem[{Jiang et~al.}(2020){Jiang, Angarita, Chiky, Cormier, \andnamec{}
  Rousseaux}]{hal-02536389}
Jiang, S., R.~Angarita, R.~Chiky, S.~Cormier, \andname{} F.~Rousseaux (2020).
\newblock {Towards the Integration of Agricultural Data From Heterogeneous
  Sources: Perspectives for the French Agricultural Context Using Semantic
  Technologies}.
\newblock In {\em {International Workshop on Information Systems Engineering
  for Smarter Life (ISESL)}}, Grenoble, France.

\bibitem[{Kingma \andnamec{} Ba}(2017){Kingma \andnamec{} Ba}]{kingma2017adam}
Kingma, D.~P. \andname{} J.~Ba (2017).
\newblock Adam: A method for stochastic optimization.

\bibitem[{Le et~al.}(2020){Le, Vial, Frej, Segonne, Coavoux, Lecouteux,
  Allauzen, Crabbé, Besacier, \andnamec{} Schwab}]{le2020flaubert}
Le, H., L.~Vial, J.~Frej, V.~Segonne, M.~Coavoux, B.~Lecouteux, A.~Allauzen,
  B.~Crabbé, L.~Besacier, \andname{} D.~Schwab (2020).
\newblock Flaubert: Unsupervised language model pre-training for french.

\bibitem[{Martin et~al.}(2020){Martin, Muller, Su{\'a}rez, Dupont, Romary,
  de~la Clergerie, Seddah, \andnamec{} Sagot}]{Martin2020CamemBERTAT}
Martin, L., B.~Muller, P.~J.~O. Su{\'a}rez, Y.~Dupont, L.~Romary, E.~de~la
  Clergerie, D.~Seddah, \andname{} B.~Sagot (2020).
\newblock Camembert: a tasty french language model.
\newblock {\em ArXiv\/}~{\em abs/1911.03894}.

\bibitem[{Pires et~al.}(2019){Pires, Schlinger, \andnamec{}
  Garrette}]{MultilingualBERT}
Pires, T., E.~Schlinger, \andname{} D.~Garrette (2019).
\newblock How multilingual is multilingual bert?
\newblock {\em CoRR\/}~{\em abs/1906.01502}.

\bibitem[{Trivedi}(2020){Trivedi}]{fast-bert}
Trivedi, K. (2020).
\newblock Fast-bert.
\newblock \url{https://github.com/kaushaltrivedi/fast-bert}.

\bibitem[{Turenne}(){Turenne}]{BSV:OCR}
Turenne, N.
\newblock reportsocr.zip.
\newblock
  \url{https://www.data.gouv.fr/fr/datasets/r/c745b0bf-b135-4dc0-ba04-1e15c1b77899}.

\bibitem[{Turenne et~al.}(2015){Turenne, Andro, CORBI{\`E}RE, \andnamec{}
  Phan}]{turenne:hal-01145489}
Turenne, N., M.~Andro, R.~CORBI{\`E}RE, \andname{} T.~T. Phan (2015).
\newblock {Open Data Platform for Knowledge Access in Plant Health Domain :
  VESPA Mining}.
\newblock working paper or preprint.

\bibitem[{Wang et~al.}(2015){Wang, Abdelzaher, \andnamec{}
  Kaplan}]{wang2015social}
Wang, D., T.~Abdelzaher, \andname{} L.~Kaplan (2015).
\newblock {\em Social sensing: building reliable systems on unreliable data}.
\newblock Morgan Kaufmann.

\bibitem[{Wolf et~al.}(2020){Wolf, Debut, Sanh, Chaumond, Delangue, Moi,
  Cistac, Rault, Louf, Funtowicz, Davison, Shleifer, von Platen, Ma, Jernite,
  Plu, Xu, Scao, Gugger, Drame, Lhoest, \andnamec{}
  Rush}]{wolf-etal-2020-transformers}
Wolf, T., L.~Debut, V.~Sanh, J.~Chaumond, C.~Delangue, A.~Moi, P.~Cistac,
  T.~Rault, R.~Louf, M.~Funtowicz, J.~Davison, S.~Shleifer, P.~von Platen,
  C.~Ma, Y.~Jernite, J.~Plu, C.~Xu, T.~L. Scao, S.~Gugger, M.~Drame, Q.~Lhoest,
  \andname{} A.~M. Rush (2020).
\newblock Transformers: State-of-the-art natural language processing.
\newblock In {\em Proceedings of the 2020 Conference on Empirical Methods in
  Natural Language Processing: System Demonstrations}, Online, pp.\  38--45.
  Association for Computational Linguistics.

\bibitem[{Zahera}(2019){Zahera}]{Zahera2019FinetunedBM}
Zahera, H.~M. (2019).
\newblock Fine-tuned bert model for multi-label tweets classification.
\newblock In {\em TREC}.

\end{thebibliography}

\end{document}